\documentclass[10pt,twocolumn,letterpaper]{article}

\usepackage{cvpr}
\usepackage{times}
\usepackage{epsfig}
\usepackage{graphicx}
\usepackage{amsmath}
\usepackage{amssymb}
\usepackage{subfigure}
\usepackage{color}
\usepackage{algorithmic} 
\usepackage{cite}
\usepackage{tabularx}
\usepackage{multirow}
\usepackage{caption2}
\usepackage{romannum}

\usepackage[pagebackref=true,breaklinks=true,letterpaper=true,colorlinks,bookmarks=false]{hyperref}

\cvprfinalcopy 

\def\cvprPaperID{807} 

\ifcvprfinal\fi
\begin{document}
\pagenumbering{arabic}
\title{Supervised Quantization for Similarity Search}

\author{Xiaojuan Wang\textsuperscript{1}\quad Ting Zhang\textsuperscript{2}
\thanks{This work was partly done when Xiaojuan Wang and Ting Zhang were interns at MSR.
They contributed equally to this work.}
\quad
Guo-Jun Qi\textsuperscript{3}\quad
Jinhui Tang\textsuperscript{4}\quad
Jingdong Wang\textsuperscript{5}\\
\textsuperscript{1}Sun Yat-sen University, China\quad
\textsuperscript{2}University of Science and Technology of China, China\\
\textsuperscript{3}University of Central Florida, USA\quad
\textsuperscript{4}Nanjing University of Science and Technology, China\\
\textsuperscript{5}Microsoft Research, China\\
}

\maketitle
\thispagestyle{empty}

\begin{abstract}
In this paper,
we address the problem of searching
for semantically similar images
from a large database. We present a compact coding approach,
supervised quantization. Our approach simultaneously learns
feature selection
that linearly transforms the database points
into a low-dimensional discriminative subspace,
and quantizes the data points in the transformed space. The optimization criterion is
that the quantized points
not only approximate the transformed points accurately,
but also are semantically separable:
the points belonging to a class
lie in a cluster that is not overlapped
with other clusters corresponding to other classes,
which is formulated as a classification problem.
The experiments on several standard datasets
show the superiority of our approach
over the state-of-the art supervised hashing and unsupervised quantization algorithms.

\end{abstract}

\section{Introduction}
Similarity search has been a fundamental research topic
in machine learning,
computer vision, and information retrieval.
The goal, given a query, is to
find the most similar item
from a database, e.g., composed of $N$ $d$-dimensional vectors.
Parallel to the study 
of indexing algorithms, 
such as kd-trees~\cite{MujaL09, WangWJLZZH13, MujaL14}, neighbhorhood graph search~\cite{WangL12},
and so on,
the recent study shows
that the compact coding approach,
including hashing and quantization,
is advantageous in terms of
memory cost,
search efficiency,
and search accuracy.


The compact coding approach
converts the database items
into short codes
in which the distance is efficiently computed.
The objective is that
the similarity computed in the coding space
is well aligned with
the similarity that is computed based on the Euclidean distance in the input space,
or that comes from the given semantic similarity
(e.g., the data items from the same class should be similar).
The solution to the former kind of similarity search
is unsupervised compact coding,
such as hashing~\cite{gionis1999similarity, jain2008fast, kulis2009kernelized,jiang2014revisiting,
xu2011complementary, gong2013iterative, kong2012isotropic, joly2011random, weiss2009spectral, weiss2012multidimensional,liu2014discrete, liu2011hashing, shen2015hashing, wang2013order, shen2013inductive, jiang2015scalable, Carreira-Perpinan2015CVPR} and quantization~\cite{jegou2011product, norouzi2013cartesian, zhang2014composite,WangWSXSL15,ZhangQTW15}.
The solution to the latter problem is supervised compact coding,
which is our interest in this paper.

Almost all research efforts in supervised compact coding
focus on developing hashing algorithms to preserve semantic similarities,
such as LDA Hashing~\cite{strecha2012ldahash},
minimal loss hashing~\cite{norouzi2011minimal},
supervised hashing with kernels~\cite{liu2012supervised},
FastHash~\cite{lin2014fast}, triplet loss hashing~\cite{norouzi2012hamming}, and supervised discrete hashing~\cite{shen2015supervised}.
In contrast,
there is less study in quantization,
which however already shows the superior performance
for Euclidean distance and cosine-based similarity search.
This paper makes a study
on the quantization solution to semantic similarity search.

Our main contributions are as follows:
(\romannum{1}) We propose a supervised composite quantization approach.
To the best of our knowledge,
our method is the first attempt to explore quantization
for semantic similarity search. The advantage of quantization over hashing is that the number of possible distances is significantly higher, and hence the distance approximation, accordingly the similarity search accuracy, is more accurate.
(\romannum{2}) Our approach jointly optimizes the quantization
and learns the discriminative subspace where the quantization is performed.
The criterion is
the semantic separability:
the points belonging to a class
lie in a cluster that is not overlapped
with other clusters corresponding to other classes,
which is formulated as a classification problem.
(\romannum{3}) Our method significantly
outperforms many state-of-the-art methods in terms of search
accuracy and search efficiency
under the same code length.

\section{Related work}

There are two main research issues in supervised hashing:
how to design hash functions
and how to preserve semantic similarity.
In essence,
most algorithms can adopt various hash functions, e.g., an algorithm using a linear hash function
usually can also use a kernel hash function.
Our review of the supervised hashing algorithms focuses on the semantic similarity preserving manners.
We roughly divide them into three categories:
pairwise similarity preserving,
multiwise similarity preserving,
and classification.

Pairwise similarity preserving hashing aligns
the similarity over each pair of items computed in the hash codes
with the semantic similarity in various manners.
Representative algorithms include LDA Hashing~\cite{strecha2012ldahash},
minimal loss hashing~\cite{norouzi2011minimal}, binary reconstructive embedding~\cite{kulis2009learning},
supervised hashing with kernels~\cite{liu2012supervised}, two-step hashing~\cite{lin2013general}, FastHash~\cite{lin2014fast}, and so on. The recent work~\cite{erin2015deep}, supervised deep hashing, designs deep neural network as hash functions to seek multiple hierarchical non-linear feature transformations, and preserves the pairwise semantic similarity by maximizing the inter-class variations and minimizing the intra-class variations of the hash codes.

Multiwise similarity preserving hashing formulates the problem
by maximizing the agreement of the similarity orders over more than two items
between the input space and the coding space.
The representative algorithms include order preserving hashing~\cite{wang2013order},
which directly aligns the rank orders computed from the input space and the coding space,
triplet loss hashing~\cite{norouzi2012hamming}, listwise supervision hashing~\cite{wang2013learning}, and so on.
Triplet loss hashing and listwise supervision hashing adopt different loss functions
to align the similarity order in the coding space and the semantic similarity over triplets of items. The recent proposed deep semantic ranking based method~\cite{zhao2015deep}  preserves multilevel semantic similarity between multilabel images by jointly learning feature representations and mappings from them to
hash codes.

The recently-developed supervised discrete hashing (SDH) algorithm~\cite{shen2015supervised}
formulates the problem
using the rule that the classification performance over the learned binary codes is as good as possible.
This rule seems inferior compared with pairwise and multiwise similarity preserving,
but yields superior search performance. This is
mainly thanks to its optimization algorithm (directly optimize the binary codes)
and scalability (not necessarily do the sampling as done in most pairwise and multiwise similarity preserving algorithms).
Semantic separability in our approach, whose goal is
that the points belonging to a class
lie in a cluster that is not overlapped
with other clusters corresponding to other classes,
is formulated as a classification problem, which can also be optimized using all the data points.

Our approach is a supervised version
of quantization.
The quantizer we adopt is  composite quantization~\cite{zhang2014composite},
which is shown to be a generalized version of
product quantization~\cite{jegou2011product} and cartesian k-means~\cite{norouzi2013cartesian},
and achieves better performance.
Rather than performing the quantization in the input space,
our approach conducts the quantization in a discriminative space,
which is jointly learned with the composite quantizer.

\section{Formulation}\label{section::formulation}

Given a $d$-dimensional query vector ${\bf q} \in \mathbb{R}^d$
and a search database
consisting of $N$ $d$-dimensional vectors
$\mathcal{X} = \{{\bf x}_{n}\}_{n = 1}^{N}$
with each point ${\bf x}_n\in \mathbb{R}^{d}$
associated with a class label, denoted by a binary label vector $\mathbf{y}_n \in \{0,1\}^C$
in which the $1$-valued entry indicates the class label of $\mathbf{x}_n$,
the goal is to find $K$ vectors from the database $\mathcal{X}$
that are nearest to the query
so that the found vectors
share the same class label with the query.
This paper is interested
in the approximate solution:
converting the database vectors into compact codes
and then performing the similarity search in the compact coding space,  which has the advantage of lower memory cost and higher search efficiency.

\vspace{.2cm}
\noindent\textbf{Modeling.}
We present a supervised quantization approach
to approximate each database vector
with a vector selected or composed from a dictionary of base items. Then the database vector is
represented by a short code composed of
the indices of the selected base items.
Our approach, rather than directly quantizing the database vectors in the original space,
learns to transform the database vectors to a discriminative subspace with a matrix $\mathbf{P} \in \mathbb{R}^{d\times r}$,
and then does the quantization in the transformed space.

We propose to adopt the state-of-the-art unsupervised quantization approach: composite quantization\cite{zhang2014composite}.
Composite quantization approximates a vector $\mathbf{x}$
using the sum of $M$ elements with each selected from a dictionary, i.e.,
$\bar{\mathbf{x}} = \sum_{m=1}^M \mathbf{c}_{mk_m}$,
where $\mathbf{c}_{mk_m}$ is selected
from the $m$th dictionary
with $K$ elements
$\mathbf{C}_m = [\mathbf{c}_{m1}~\mathbf{c}_{m2}~\cdots~\mathbf{c}_{mK}]$,
and encodes $\mathbf{x}$
by a short code $(k_1~k_2~\cdots~k_M)$.
Our approach uses the sum to approximate the transformed vector,
which is formulated by minimizing the approximation error,
\begin{align}
\|\mathbf{P}^T\mathbf{x} - \bar{\mathbf{x}}\|_{2}^{2} = \|\mathbf{P}^T\mathbf{x} - \sum\nolimits_{m=1}^M \mathbf{c}_{mk_m} \|_2^2.
\end{align}

We learn the transformation matrix $\mathbf{P}$
such that the quantized data points
are semantically separable:
the points belonging to the same class
lie in a cluster,
and the clusters corresponding to different classes
are disjointed.
We solve the semantic separation problem by finding $C$ linear decision surfaces to divide all the points into $C$ clusters\footnote{$C$ linear decision surfaces can divide the points into more than $C$ clusters.},
each corresponding to a class,
which is formulated
as a classification problem
given as follows,
\begin{align}
\sum\nolimits_{n=1}^N\ell(\mathbf{y}_n, \mathbf{W}^T\bar{\mathbf{x}}_n) + \lambda \|\mathbf{W}\|_F^2,
\end{align}
where $\lambda$ is the parameter controlling
the regularization term $\|\mathbf{W}\|_F^2$;
$\mathbf{W} = [\mathbf{w}_1~\mathbf{w}_2~\cdots~\mathbf{w}_C] \in \mathbb{R}^{r \times C}$;
$\ell(\cdot , \cdot)$ is a classification loss function
to penalize the cases where the point $\bar{\mathbf{x}}_n$
is not assigned to the cluster corresponding to $\mathbf{y}_n$
based on the $C$ associated decision functions $\{\mathbf{w}_k^T\bar{\mathbf{x}}_n\}_{k=1}^C$.
In this paper, we adopt the regression loss:
\begin{align}
	\ell(\mathbf{y}_n, \mathbf{W}^T\bar{\mathbf{x}}_n) = \|\mathbf{y}_n - \mathbf{W}^T\bar{\mathbf{x}}_n\|_2^2
\end{align}

The proposed approach combines the quantization with the feature selection,
and jointly learns the quantization parameter and the transform matrix.
The overall objective function is given as follows,
\begin{align}
\min_{\mathbf{W}, \mathbf{P}, \mathbf{C}, \{\mathbf{b}_n\}_{n=1}^N, \epsilon}~&~\sum\nolimits_{n=1}^N \|\mathbf{y}_n - \mathbf{W}^T\mathbf{C}\mathbf{b}_n\|_2^2
+ \lambda \|\mathbf{W}\|_F^2 \nonumber \\
~&~+\gamma \sum\nolimits_{n=1}^N \|\mathbf{C}\mathbf{b}_n - \mathbf{P}^T\mathbf{x}_n\|_2^2  \label{equation::obj}  \\
\operatorname{s.t.}~&~\sum\nolimits_{i \neq j}^M \mathbf{b}^T_{ni}\mathbf{C}^T_i\mathbf{C}_j\mathbf{b}_{nj} = \epsilon, \nonumber\\
 ~&~n=1,2,\cdots, N,\nonumber
\end{align}
where $\gamma$ is the parameter controlling the quantization term; $\mathbf{C}\mathbf{b}_n$ is the matrix form of
$\sum_{m=1}^M\mathbf{c}_{mk_m^n}$
and $\mathbf{b}_n = [\mathbf{b}_{n1}^T~\mathbf{b}_{n2}^T~\cdots~\mathbf{b}_{nM}^T]^T$;
$\mathbf{b}_{nm} \in \{0, 1\}^K$ is an indicator vector
with only one entry being $1$,
indicating that the corresponding dictionary element is selected
from the $m$th dictionary.
The equality constraint,
$ \sum_{i \neq j}^M \mathbf{b}^T_{ni}\mathbf{C}^T_i\mathbf{C}_j\mathbf{b}_{nj}
= \sum_{i \neq j}^M \mathbf{c}^T_{ik^n_i} \mathbf{c}_{jk^n_j} =\epsilon$,
called constant inter-dictionary-element-product,
is introduced from composite quantization~\cite{zhang2014composite}
for fast distance computation (reduced from $O(d)$ to $O(M)$) in the search stage, which is presented below.

\vspace{.2cm}
\noindent{\bf Querying.}
The search process is similar to that in composite quantization~\cite{zhang2014composite}.
Given a query ${\bf q}$, after transformation,
the approximate distance
between ${\bf q}$ (represented as $\mathbf{q}' = \mathbf{P}^T\mathbf{q}$)
and a database vector ${\bf x}$
(represented
as $\mathbf{C}\mathbf{b} = \sum_{m=1}^M \mathbf{c}_{mk_m}$)
is computed as
	\begin{align}
		~&~ \| \mathbf{q}' - \sum_{m=1}^M \mathbf{c}_{mk_m}\|_2^2 = 	\label{eqn:distanceapproximation} \\
		 ~&~ \sum_{m=1}^M \|\mathbf{q}' - \mathbf{c}_{mk_m}\|_2^2
		- (M-1)\|\mathbf{q}'\|_2^2 + \sum_{i \neq j}^M  \mathbf{c}^T_{ik_i} \mathbf{c}_{jk_j}.\nonumber
	\end{align}
	
Given the query $\mathbf{q}'$, the second term $-(M-1)\|\mathbf{q}'\|_2^2$ in the right-hand side of Equation~\ref{eqn:distanceapproximation} is a constant for all database vectors. Meanwhile, the third term $\sum_{i \neq j}^M  \mathbf{c}^T_{ik_i} \mathbf{c}_{jk_j}$, which is equal to $\epsilon$ thanks to the introduced constant constraint, is also a constant. Hence these two constant terms  can be ignored,
as they do not affect the sorting results.
As a result,
it is enough to compute
the distances between $\mathbf{q}'$
and the selected dictionary elements $\{\mathbf{c}_{mk_m}\}_{m=1}^M$: $\{\|\mathbf{q}' - \mathbf{c}_{mk_m}\|_2^2\}_{m=1}^M$.
We precompute a distance table of length $MK$
recording the distances between $\mathbf{q}'$
and the dictionary elements in all the dictionaries
before examining the distance between $\mathbf{q}'$
and each approximated point $\bar{\mathbf{x}}$ in the database.
Then computing $\sum_{m=1}^M \|\mathbf{q}' - \mathbf{c}_{mk_m}\|_2^2$ takes
only $O(M)$ distance table lookups and $O(M)$ addition operations.

\section{Optimization}\label{section::optimization}

Our problem~(\ref{equation::obj})
consists of five groups of unknown variables:
classification matrix ${\bf W}$, transformation matrix ${\bf P}$, dictionaries ${\bf C}$, binary indicator vectors $\{{\bf b}_n\}_{n = 1}^{N}$,
and the constant $\epsilon$.
We follow~\cite{zhang2014composite} and combine the constraints $\sum_{i \neq j}^M{\bf b}_{ni}^T{\bf C}_i^T{\bf C}_j{\bf b}_{nj} = \epsilon$
into the objective function using the quadratic penalty method:
\begin{equation} \small
\begin{aligned}
	~&~\psi({\bf W},{\bf P}, {\bf C}, \{{\bf b}_n\}_{n = 1}^{N}, \epsilon) = \sum_{n=1}^N \|\mathbf{y}_n - \mathbf{W}^T\mathbf{C}\mathbf{b}_n\|_2^2
	+ \lambda \|\mathbf{W}\|_F^2 \\
	~&~+\gamma\sum_{n=1}^N \|\mathbf{C}\mathbf{b}_n - \mathbf{P}^T\mathbf{x}_n\|_2^2  +\mu \sum_{n = 1}^{N}(\sum_{i\neq j}^M{\bf b}_{ni}^T{\bf C}_i^T{\bf C}_j{\bf b}_{nj} - \epsilon)^2,
\end{aligned}
\label{equation::objToBeOptimized}
\end{equation}
where $\mu$ is the penalty parameter.

We use the alternative optimization technique
to iteratively solve the problem,
with each iteration updating one of ${\bf W}, {\bf P}, \epsilon, {\bf C}$, and $\{{\bf b}_n\}_{n = 1}^{N}$
while fixing the others. The initialization scheme
and the iteration details are presented as follows.

\vspace{0.2cm}
\noindent {\bf Initialization.}
The transformation matrix  ${\bf P}$ is initialized using principal component analysis (PCA).
We use the dictionaries and codes learned from product quantization~\cite{jegou2011product}
in the transformed space to initialize ${\bf C}$ and $\{{\bf b}_n\}_{n = 1}^{N}$ for the shortest code ($16$ bits) in our experiment,
and we use the dictionaries and codes learned in the shorter code to
do the initialization for longer code
with setting the additional dictionary elements to zero and randomly initializing the additional binary codes.

\vspace{0.2cm}
\noindent {\bf ${\bf W}$-Step.} With ${\bf C}$ and $\{{\bf b}_n\}_{n = 1}^{N}$ fixed, ${\bf W}$ is solved by the regularized least squares problem, resulting in a closed-form solution:
\begin{equation}
	{\bf W}^* = (\bar{\bf X}\bar{\bf X}^T + \lambda {\bf I}_r)^{-1}\bar{\bf X}{\bf Y}^T,
\end{equation}
where $\bar{\bf X} = [{\bf C}{\bf b}_1~\cdots~{\bf C}{\bf b}_N]\in \mathbb{R}^{r\times N}$, $\mathbf{Y} = [{\bf y}_1~\cdots~{\bf y}_N]\in \mathbb{R}^{C\times N}$, and $\mathbf{I}_r$ is an identity matrix of size $r \times r$.

\vspace{0.2cm}
\noindent {\bf $\bf P$-Step.} With ${\bf C}$ and $\{{\bf b}_n\}_{n = 1}^{N}$ fixed, the transformation matrix $\mathbf{P}$ is solved using the normal equation:
\begin{equation}
	{\bf P}^* = ({\bf X}{\bf X}^T)^{-1}{\bf X}\bar{\bf X}^T
	\label{equation::solution of p},
\end{equation}
where ${\bf X} = [{\bf x}_1~\cdots~{\bf x}_N]\in \mathbb{R}^{d\times N}$. 

\vspace{0.2cm}
\noindent {\bf $\epsilon$-Step.} With ${\bf C}$ and $\{{\bf b}_n\}_{n = 1}^{N}$ fixed, the objective function is a quadratic function with respect to $\epsilon$, and it is easy to get the optimal solution to $\epsilon$.
\begin{equation}
	\epsilon^* = \frac{1}{N}\sum_{n = 1}^{N}\sum_{i \neq j}^{M}{\bf b}_{ni}^T{\bf C}_i^T{\bf C}_j{\bf b}_{nj}.
	\label{equation::solution of epsilon}
\end{equation}

\vspace{0.2cm}
\noindent {\bf $\bf C$-Step.} With other variables fixed, the problem is an unconstrained nonlinear optimization problem with respect to ${\bf C}$.
We use the quasi-Newton algorithm and specifically the L-BFGS algorithm, the limited version of the Broyden-Fletcher-Goldfarb-Shanno (BFGS) algorithm. The implementation is publicly available\footnote{http://users.iems.northwestern.edu/~nocedal/lbfgsb.html}.
The derivative with respect to ${\bf C}$ and the objective function value need to be fed into the solver. L-BFGS is an iterative algorithm and we set its maximum iterations to $100$.
The partial derivative with respect to ${\bf C}_m$ is :
	\begin{align}
\frac{\partial \psi}{\partial{\bf C}_m}=~&
		\sum_{n = 1}^{N}[2{\bf W}({\bf W}^T{\bf C}{\bf b}_n - {\bf y}_n){\bf b}_{nm}^T+\\
		& 2\gamma({\bf C}{\bf b}_{n}-{\bf P}^T{\bf x}_{n}){\bf b}_{nm}^T + \nonumber \\
		& 4\mu(\sum_{i \neq j}^{M}{\bf b}_{ni}^T{\bf C}_i^T{\bf C}_j{\bf b}_{nj} -\epsilon)(\sum_{l=1,l\neq m}^{M}{\bf C}_l{\bf b}_{nl}){\bf b}_{nm}^T]. \nonumber
			\label{equation: gradientC}
	\end{align}

\vspace{0.2cm}
\noindent {\bf B-Step.} The optimization problem with respect to $\{{\bf b}_n\}_{n = 1}^{N}$ could be decomposed to $N$ subproblems,
	\begin{align}
		&\psi_n({\bf b}_n) = ||{\bf y}_n - {\bf W}^T{\bf C}{\bf b}_n||_2^2 + \gamma||{\bf C}{\bf b}_n - {\bf P}^T{\bf x}_n||_2^2 \nonumber \\
		&+ \mu (\sum_{i\neq j}^{M}{\bf b}_{ni}^T{\bf C}_i^T{\bf C}_j{\bf b}_{nj} - \epsilon)^2.
	\end{align}
	${\bf b}_n$ is a binary-integer-mixed vector, and thus the optimization is NP-hard. We use the alternative optimization technique again to solve the $M$ subvectors $\{{\bf b}_{nm}\}_{m=1}^M$ iteratively.
With $\{{\bf b}_{nl}\}_{l=1,l \neq m}^M$ fixed,
we exhaustively check all the elements in the dictionary ${\bf C}_m$, finding the element such that $\psi_n({\bf b}_n)$ is minimized, and accordingly set the corresponding entry of ${\bf b}_{nm}$ to be $1$ and all the others to be $0$.

\begin{figure}[t]
	\centering
	\subfigure[NUS-WIDE]{\label{figures::convergenCurve for nuswide}\includegraphics[width=.49\linewidth, clip]{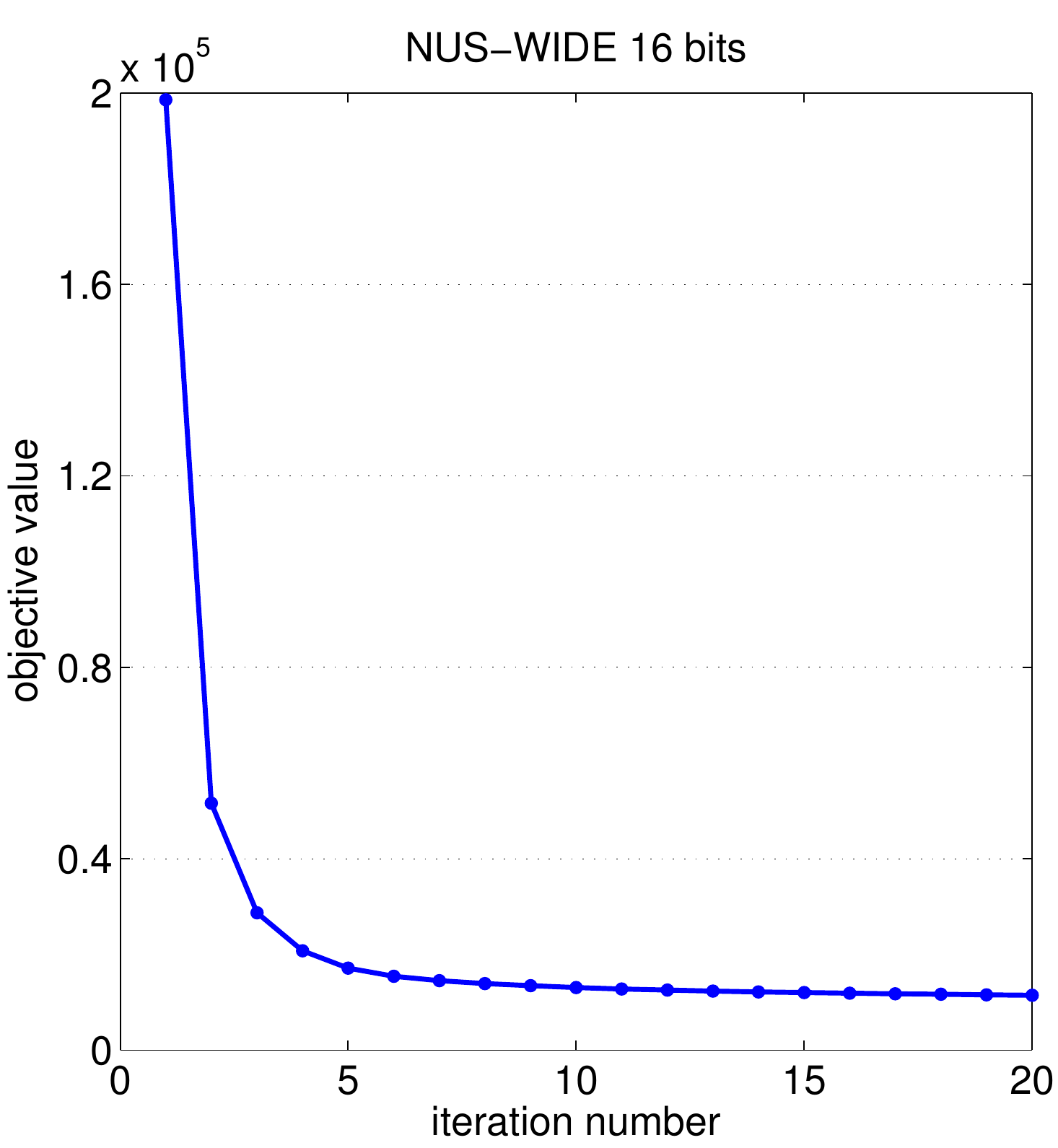}}
	\subfigure[ImageNet]{\label{figures::convergenCurve for imagenet}\includegraphics[width=.49\linewidth, clip]{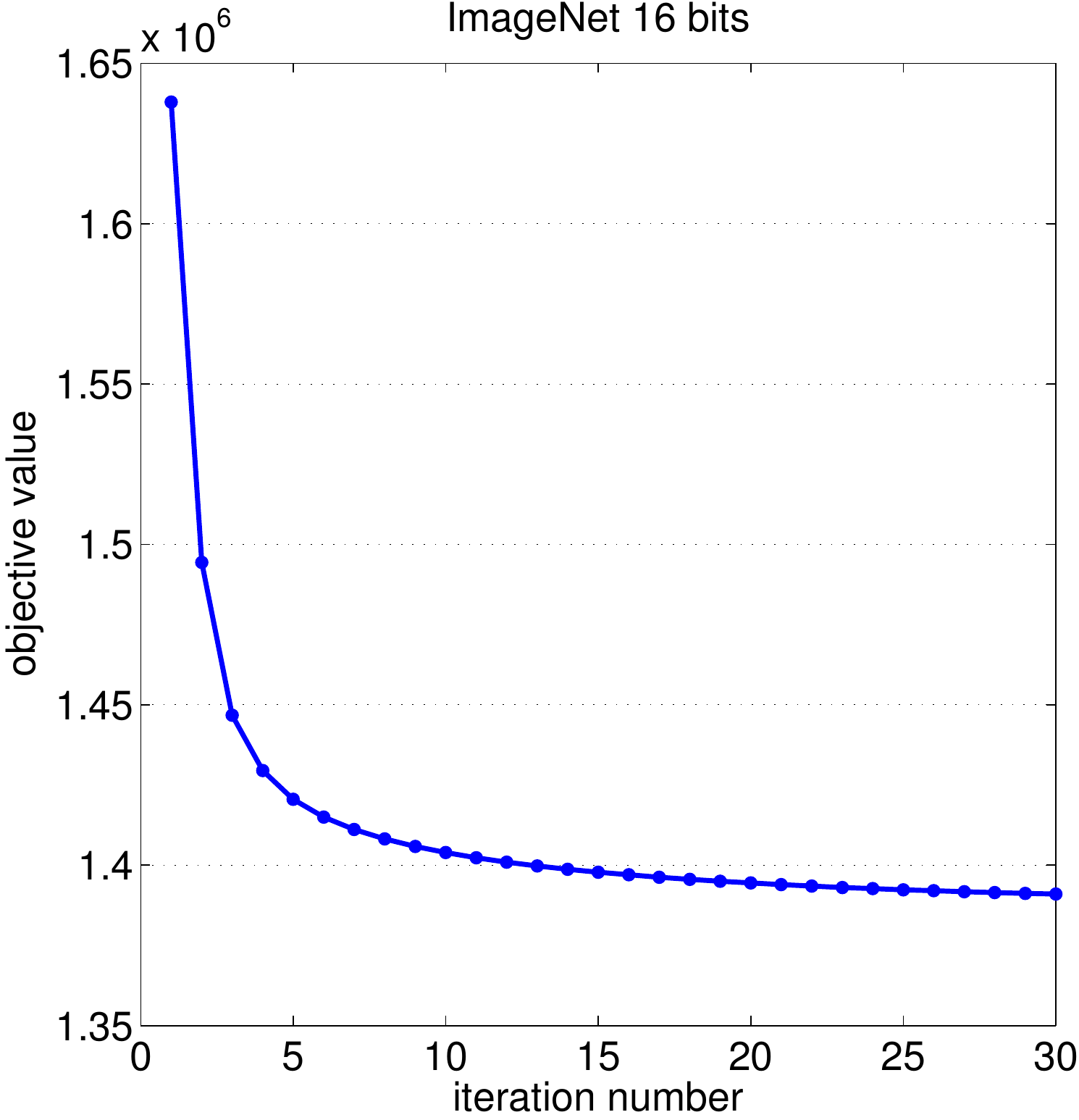}}
	\caption{Convergence curves of our algorithm on NUS-WIDE and ImageNet with 16 bits. The vertical axis represents the value of the objective function~(\ref{equation::objToBeOptimized}) and the horizontal axis corresponds to the number of iterations.}
	\label{figure::convergeCurve}
\vspace{-.5cm}
\end{figure}

\vspace{0.2cm}
\noindent {\bf Convergence.} Every update step in the algorithm assures that
the objective function value weakly decreases after each iteration, and the empirical results show that the algorithm takes a few iterations to converge.
Figure~\ref{figure::convergeCurve} shows the convergence curves on NUS-WIDE and ImageNet with $16$ bits,
which indicates that our algorithm gets converged in a few iterations.

\section{Discussions}
\noindent{\bf Connection with supervised sparse coding.}
It is pointed in~\cite{zhang2014composite}
that composite quantization
is related to sparse coding:
the binary indicator vector $\mathbf{b}$
is a special sparse code,
containing only $M$ non-zero entries (valued as $1$)
and each non-zero entry distributed in a subvector.
The proposed supervised quantization approach
is close to supervised sparse coding~\cite{MairalBPSZ08},
which introduces supervision to learn the dictionary and the sparse codes,
but different from it
in the motivation and the manner of imposing the supervision:
our approach adopts the supervision to help separate the data points
into clusters with each corresponding to a class;
our approach imposes the supervision
on the approximated data points
while supervised sparse coding imposes the supervision
on the sparse codes.

\vspace{0.2cm}
\noindent {\bf Classification loss vs. rank loss.}
There are some hashing approaches
exploring the supervision information
through rank loss~\cite{WangSSJ14},
such as
the triplet loss in~\cite{wang2013learning, norouzi2012hamming},
and the pairwise loss in~\cite{wang2012semi, norouzi2011minimal}.
In general, compared with the classification loss,
those two rank losses might be more helpful to learn the compact codes
as they directly align the rank order in the coding space
with the given semantic rank information.
However, they yield a larger number of loss terms,
e.g., $O(N^2)$ for pairwise loss
and $O(N^3)$ for triplet loss,
requiring prohibitive computational cost
which makes the optimization difficult and infeasible.
Therefore, sampling is usually adopted for training,
which however makes the results
not as good as expected. A comparison with triplet loss is shown in Section~\ref{section::emperical analysis}.

\section{Experiment}

\subsection{Datasets and settings}\label{section::datasets}

\noindent {\bf Datasets.} We perform the experiments on four standard datasets: CIFAR-10~\cite{krizhevsky2009learning},
MNIST~\cite{lecun1998gradient},
NUS-WIDE~\cite{chua2009nus},
and ImageNet\cite{deng2009imagenet}.

The CIFAR-10 dataset consists of $60,000$ $32 \times 32$ color tinny images,
and includes $10$ classes with $6,000$ images per class. We represent each image by a $512$-dimensional GIST feature vector available on the website\footnote{http://www.cs.toronto.edu/~kriz/cifar.html}. The dataset is split into a query set with $1,000$ samples
and a training set with all the remaining samples as done in~\cite{shen2015supervised}.

The MNIST dataset consists of $70,000$ $28 \times 28$ greyscale images of handwritten digits
from '$0$' to '$9$'.
Each image is represented
by the raw pixel values,
resulting in a 784-dimensional vector.
We split the dataset into a query set with $1,000$ samples
and a training set with all remaining samples as done in~\cite{shen2015supervised}.

The NUS-WIDE dataset contains $269,648$ images
collected from Flickr,
with each image containing multiple semantic labels from  $81$ concept labels.
The $500$-dimensional bag-of-words features provided in~\cite{chua2009nus}
are used.
Following~\cite{shen2015supervised},
we collect 193,752 images that are from the $21$ most frequent labels for evaluation, including {\em sky, clouds, person, water, animal, grass, building, window, plants, lake, ocean, road, flowers, sunset, relocation, rocks, vehicles, snow, tree, beach, }and {\em mountain}.
For each label, $100$ images are uniformly sampled as the query set,
and the remaining images are used as the training set.

The dataset ILSVRC 2012~\cite{deng2009imagenet}, named as ImageNet in this paper,
contains
over $1.2$ million images of $1,000$ categories.
We use the provided training set as the retrieval database and
the provided $50,000$ validation images as the query set
since the ground-truth labeling of the test set is not publicly available.
Similar to~\cite{shen2015supervised},
we use the $4096$-dimensional feature extracted from the convolution neural networks (CNN) in ~\cite{krizhevsky2012imagenet} to represent each image.

\vspace{0.2cm}
\noindent {\bf Evaluation criteria.}
We adopt the widely used mean average precision (MAP) criterion, defined as MAP $= \frac{1}{Q}\sum_{i = 1}^{Q}AP({\bf q}_i),$ where $Q$ is the number of queries,
and $AP$ is computed as $AP({\bf q}) = \frac{1}{L}\sum_{r = 1}^{R}P_{{\bf q}}(r)\delta(r).$
Here $L$ is the number of true neighbors for the query ${\bf q}$
in the $R$ retrieved items,
where $R$ is the size of the database
except that $R$ is $1500$
on the ImageNet dataset
for evaluation efficiency.
$P_{{\bf q}}(r)$ denotes the precision
when top $r$ data points are returned,
and $\delta(r)$ is an indicator function
which is $1$ when the $r$th result is a true neighbor and otherwise $0$.
A data point is considered as a true neighbor
when it shares at least one class label with the query.

Besides the search accuracy,
we also report the search efficiency by evaluating the query time
under various code lengths.
The query time contains the query preprocessing time and the linear scan search time. For hashing algorithms, the query preprocessing time refers to query encoding time; for unsupervised quantization algorithms, the query preprocessing time refers to distance lookup table construction time; for our proposed method, the query preprocessing time includes feature transformation time and distance lookup table construction time. For all methods, we use C++ implementations to test the query time on a $64$-bit windows server with $48$ GB RAM and $3.33$ GHz CPU.

\vspace{0.2cm}
\noindent {\bf Parameter settings.}
There are three trade-off parameters
in the objective function~(\ref{equation::objToBeOptimized}):
$\gamma$ for the quantization loss term,
$\mu$ for penalizing the equality constraint term,
and $\lambda$ for the regularization term.
We select $\gamma$ and $\mu$ via validation. We choose a subset of the training set as the validation set
(the size of the validation set is the same to that of the query set),
and
the best parameters $\gamma$ and $\mu$ are chosen
so that the average search performance in terms of MAP,
by regarding the validation vectors as queries, is the best.
It is feasible that the validation set is a subset of the training set,
as the validation criterion is not the objective function but the search performance~\cite{zhang2014composite}.
The empirical analysis about the two parameters
will be given in Section~\ref{section:: discussion of paramters}.
The parameter $\lambda$ is set to $1$,
which already shows the satisfactory performance.
We set the dimension $r$ of the discriminative subspace to $256$.
We do not tune $r$ and $\lambda$ for saving time
while we think that tuning it might yield better performance. We choose $K = 256$ to be the dictionary size as done in \cite{jegou2011product, norouzi2013cartesian, zhang2014composite}, so that the resulting distance lookup tables are small and each subindex fits into one byte.

\subsection{Comparison}
\noindent\textbf{Methods.}
Our method, denoted by SQ,
is compared with seven state-of-the-art supervised hashing methods:
supervised discrete hashing (SDH)~\cite{shen2015supervised},
FastHash\cite{wang2015fast},
supervised hashing with kernels (KSH)~\cite{liu2012supervised},
CCA-ITQ\cite{gong2013iterative},
semi-supervised hashing (SSH)~\cite{wang2012semi},
minimal loss hashing (MLH)~\cite{norouzi2011minimal},
and binary reconstructive embedding (BRE)~\cite{kulis2009learning},
as well as the state-of-the-art unsupervised quantization method, composite quantization (CQ)~\cite{zhang2014composite}. To the best of our knowledge, there do not exist supervised quantization algorithms.
We use the public implementations
for all the algorithms
except that we implement SSH by ourselves
as we do not find the public code,
and follow the corresponding papers/authors to set up the parameters. For FastHash, we adopt hinge loss as loss function in the binary code inference step and boosted tree as classifier in the hash function learning step, which is suggested by the author to achieve the best performance.

\begin{figure*}[t]
	\centering
	\subfigure[CIFAR-10]{\label{fig:ResultsOfMNIST:a}\includegraphics[width=.33\linewidth, clip]{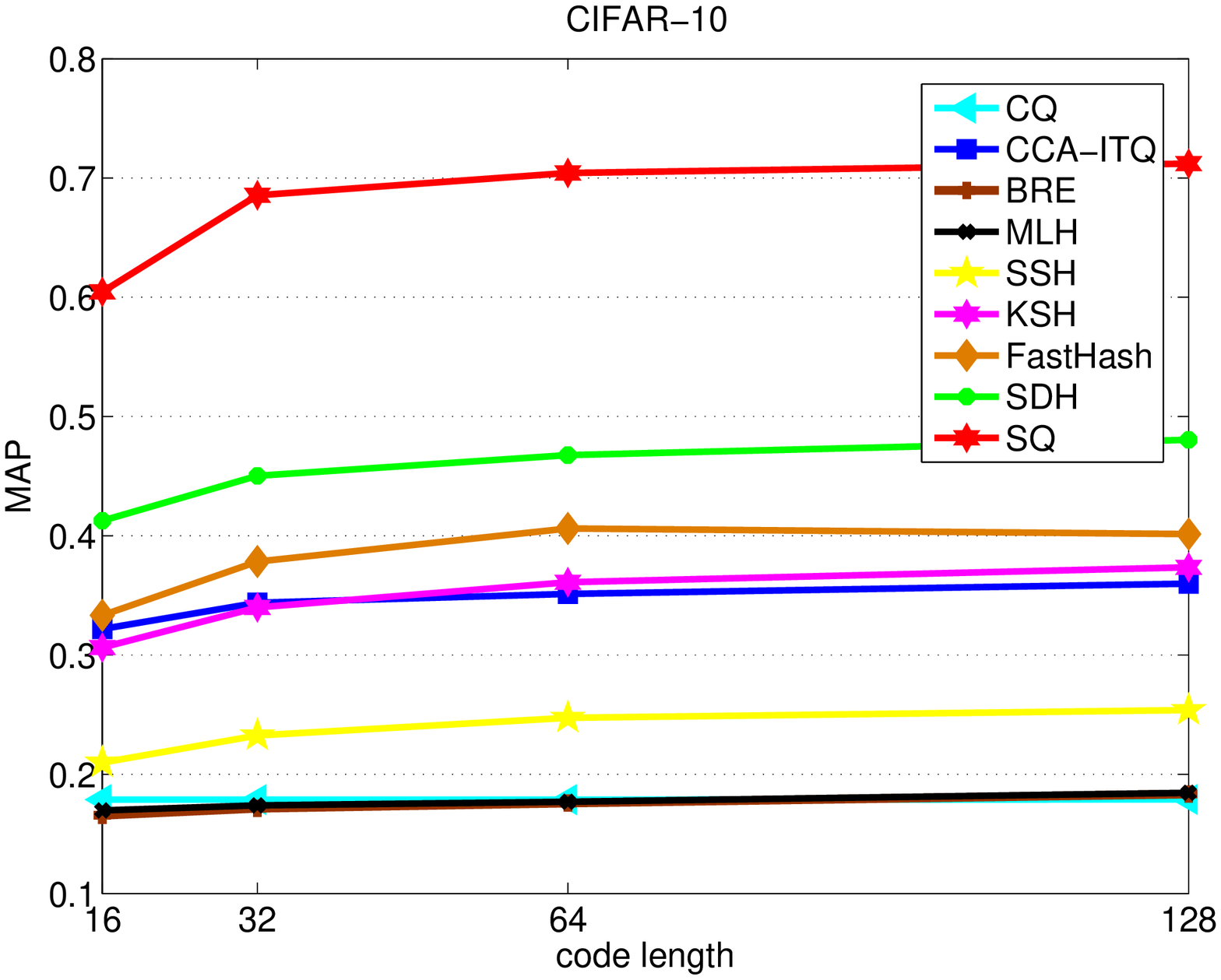}}
	\subfigure[MNIST]{\label{fig:ResultsOfMNIST:b}\includegraphics[width=.33\linewidth, clip]{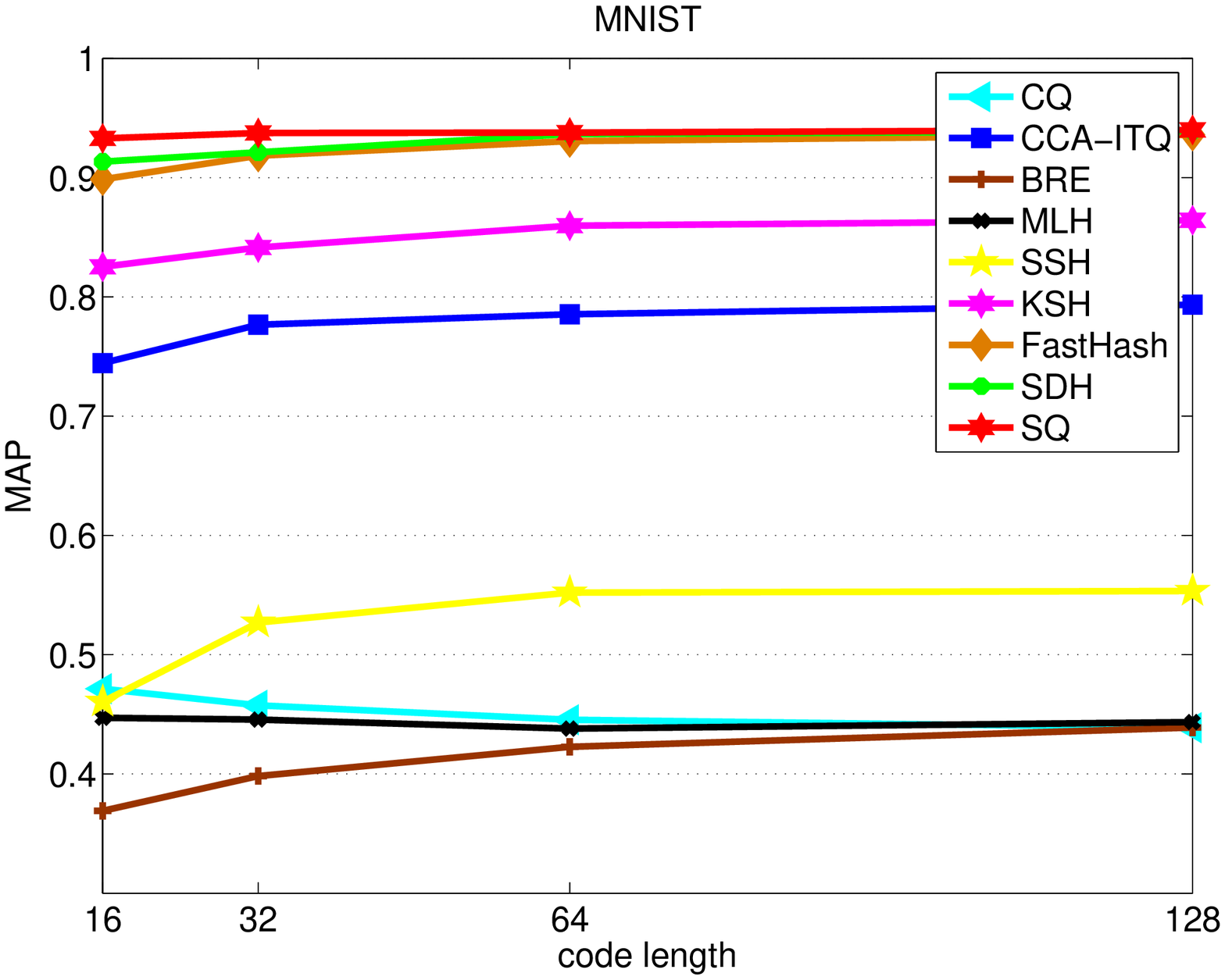}}
	\subfigure[NUS-WIDE]{\label{fig:ResultsOfMNIST:c}\includegraphics[width=.33\linewidth, clip]{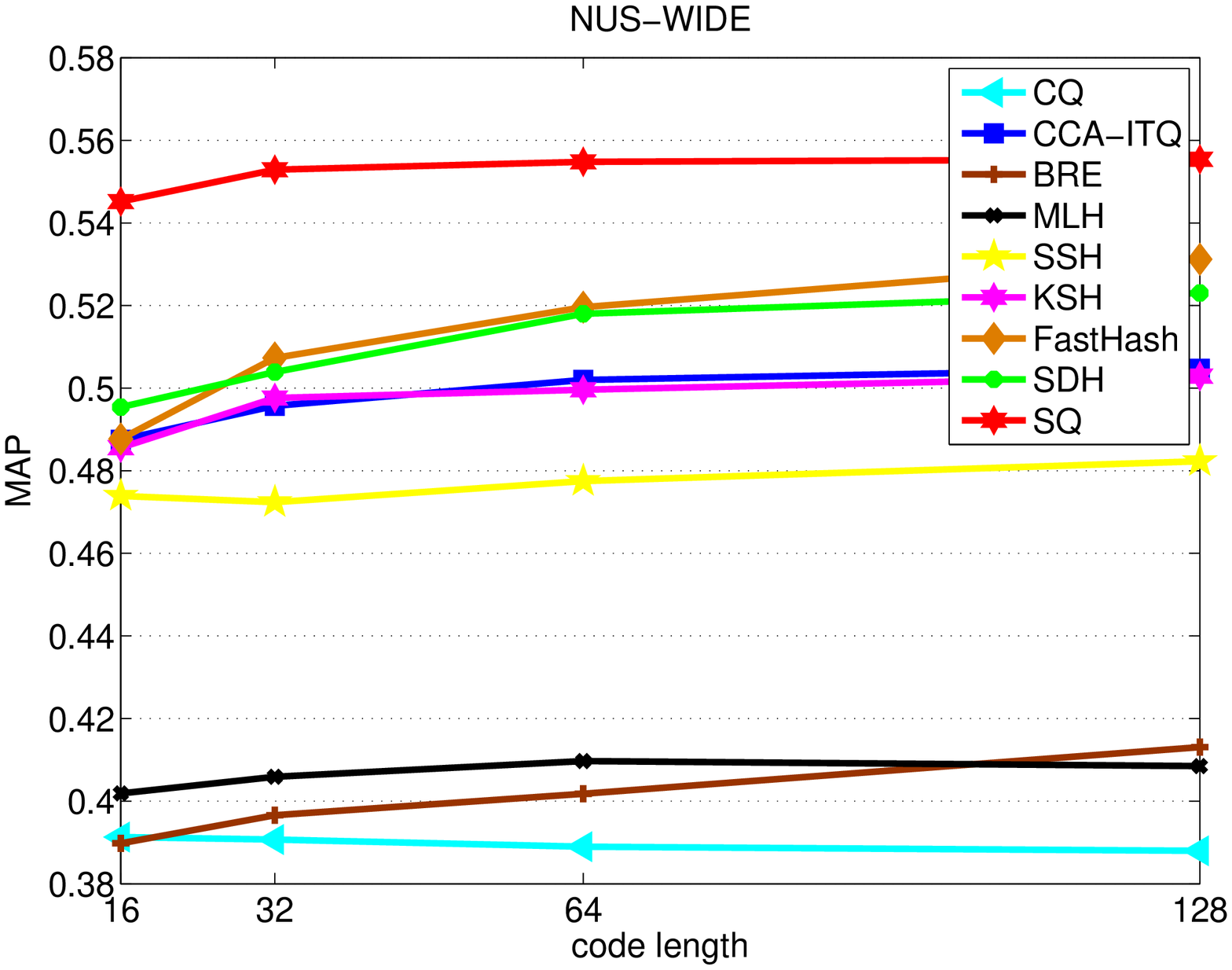}}
	\caption{Search performance (in terms of MAP) comparison  of different methods on CIFAR-10, MNIST, and NUS-WIDE with code length of $16$, $32$, $64$, and $128$.}
	\label{figure::comparison with related methods}
\vspace{-.5cm}
\end{figure*}

\vspace{0.2cm}
\noindent\textbf{Implementation details.} It is infeasible to do the training
over the whole training set
for the pairwise-similarity-based hashing algorithms (SSH, BRE, MLH, KSH, FastHash), as discussed in~\cite{shen2015supervised}.
Therefore, for CIFAR-10, MNIST, and NUS-WIDE,
following the recent work~\cite{shen2015supervised},
we randomly sample $5000$ data points from the training set to do the optimization for the pairwise similarity-based algorithms,
and use the whole training set for SDH and CCA-ITQ.
For ImageNet, we use
as many training samples for optimization
as possible
if the $256$G RAM in our server is enough for optimization:
$500,000$ for CCA-ITQ, $100,000$ for SDH, $10,000$ for the remaining hashing methods.
There are two hashing algorithms, KSH and SDH, that adopt the kernel-based representation, i.e.,
select $h$ anchor points $\{{\bf a}_j\}_{j = 1}^{h}$
and use
$\phi({\bf x})  = [\exp(-||{\bf x} - {\bf a}_1||_{2}^2/2\sigma^{2})~\dots~\exp(-||{\bf x} - {\bf a}_h||_{2}^2/2\sigma^{2})]^T \in \mathbb{R}^h$
to represent $\mathbf{x}$. Our approach also uses the kernel-based representation
for CIFAR-10, MNIST, and NUS-WIDE.
Following~\cite{shen2015supervised},
$h= 1000$
and $\sigma$ is chosen based on the rule $\sigma = \frac{1}{N}\sum_{n = 1}^N \min \{\|{\bf x}_n - {\bf a}_j\|_2\}_{j = 1}^h$.

\vspace{0.2cm}
\noindent\textbf{Search accuracy.}
The results on CIFAR-10, MNIST, and NUS-WIDE
with the code length of $16$, $32$, $64$, and $128$,
are shown in Figure~\ref{figure::comparison with related methods}.
It can be seen that
our approach, SQ, achieves the best performance,
and SDH is the second best.
In comparison with SDH,
our approach gains large improvement
on CIFAR-10 and NUS-WIDE, e.g., $23.66\%$ improvement on CIFAR-10 with $64$ bits,
and $4.65\%$ improvement on NUS-WIDE with $16$ bits.
It is worth noting that on these two datasets, the performance of SQ with $16$ bits is even much better than that of SDH with $128$ bits.
Our approach gets relatively small improvement
over SDH on MNIST. The reason might be that SDH
already achieves a high performance, and it is not easy to get a large improvement further.
Compared with the unsupervised quantization algorithm, composite quantization (CQ), whose performance  is lower than most of the supervised hashing algorithms,
our approach obtains significant improvement, e.g., $42.57\%$ improvement on CIFAR-10 with $16$ bits,  $46.14\%$ on MNIST with $16$ bits, and $15.39\%$ on NUS-WIDE with $16$ bits.
This shows that learning with supervision indeed benefits the search performance.

\begin{figure}[t]
	\centering
	\includegraphics[width= \linewidth, clip]{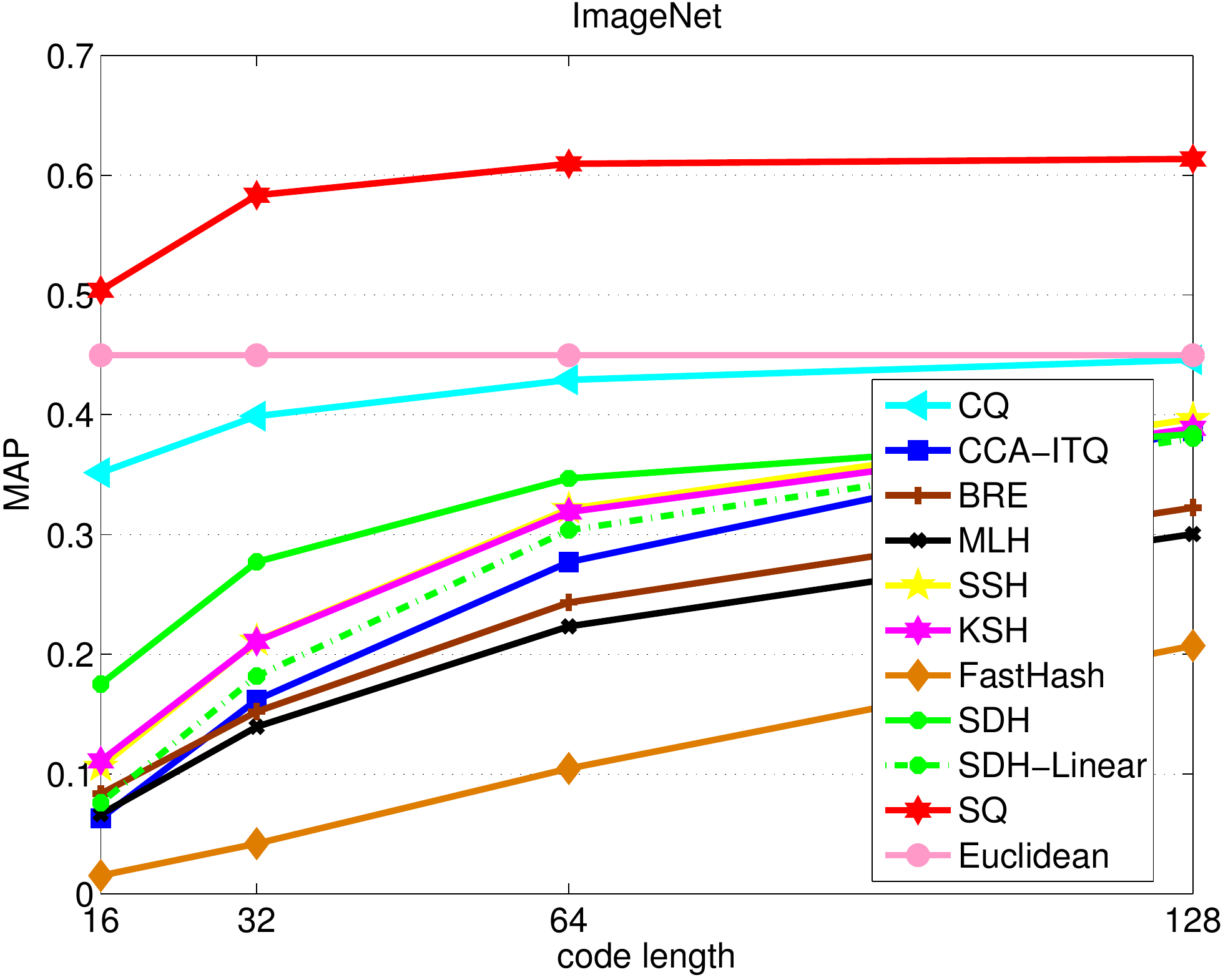}~
	\caption{Search performance (in terms of MAP) comparison  of different methods on ImageNet with code length of $16$, $32$, $64$, and $128$.}
	\label{figure::comparison with related methods in imagenet}
\vspace{-.5cm}
\end{figure}

The result on ImageNet
is shown in Figure~\ref{figure::comparison with related methods in imagenet}.
The performance of our approach again outperforms
other algorithms, and CQ is the second best. The reason might be the powerful discrimination ability of the original CNN features. To achieve a comprehensive analysis, we provide the Euclidean baseline (see Figure~\ref{figure::comparison with related methods in imagenet}) that simply computes the distances between the query and the database vectors using the original CNN features and returns the top $R$ retrieved items.  As shown in Figure \ref{figure::comparison with related methods in imagenet}, our proposed SQ also outperforms the Euclidean baseline by a large margin, and CQ is a little lower than the baseline. This shows that our approach is able to learn better quantizer through the supervision though it is known that the CNN features are already good. The best supervised hashing algorithm, SDH, uses the kernel-based representation in our experiment as suggested in its original paper~\cite{shen2015supervised}. To further verify the superiority of our approach over SDH, we also report the result of SDH without using the kernel representation (denoted by ``SDH-Linear" in Figure \ref{figure::comparison with related methods in imagenet}), and find that it is still lower than our approach. This further shows the effectiveness of quantization: quantization has much more different differences compared with hashing, which has only a few Hamming distances for the same code length.

\begin{figure*}[t]
	\centering
	\subfigure[CIFAR-10]{\label{figure:querytime_cifar10}\includegraphics[width=.24\linewidth, clip]{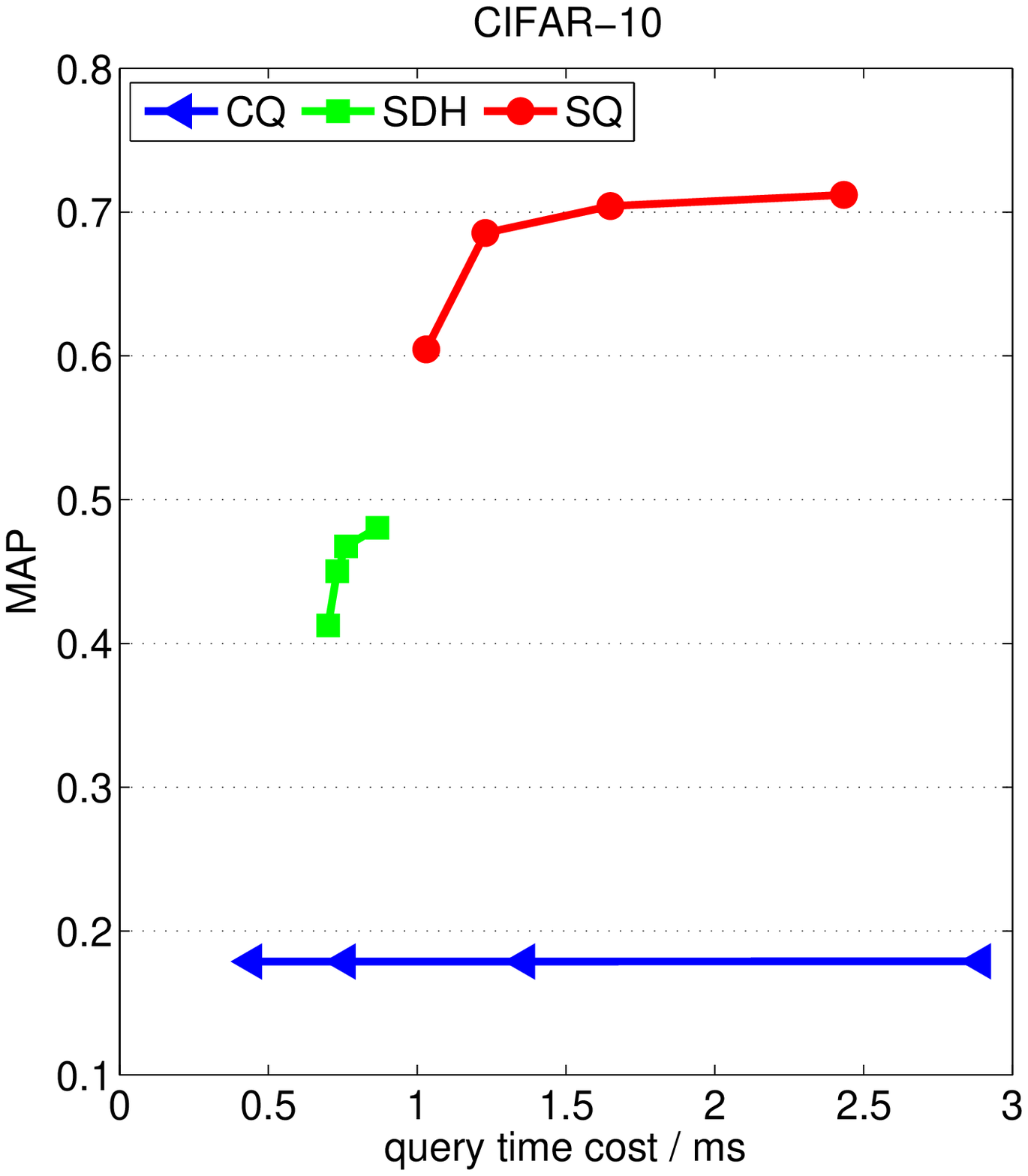}}
	\subfigure[MNIST]{\label{figure:querytime_mnist}\includegraphics[width=.24\linewidth, clip]{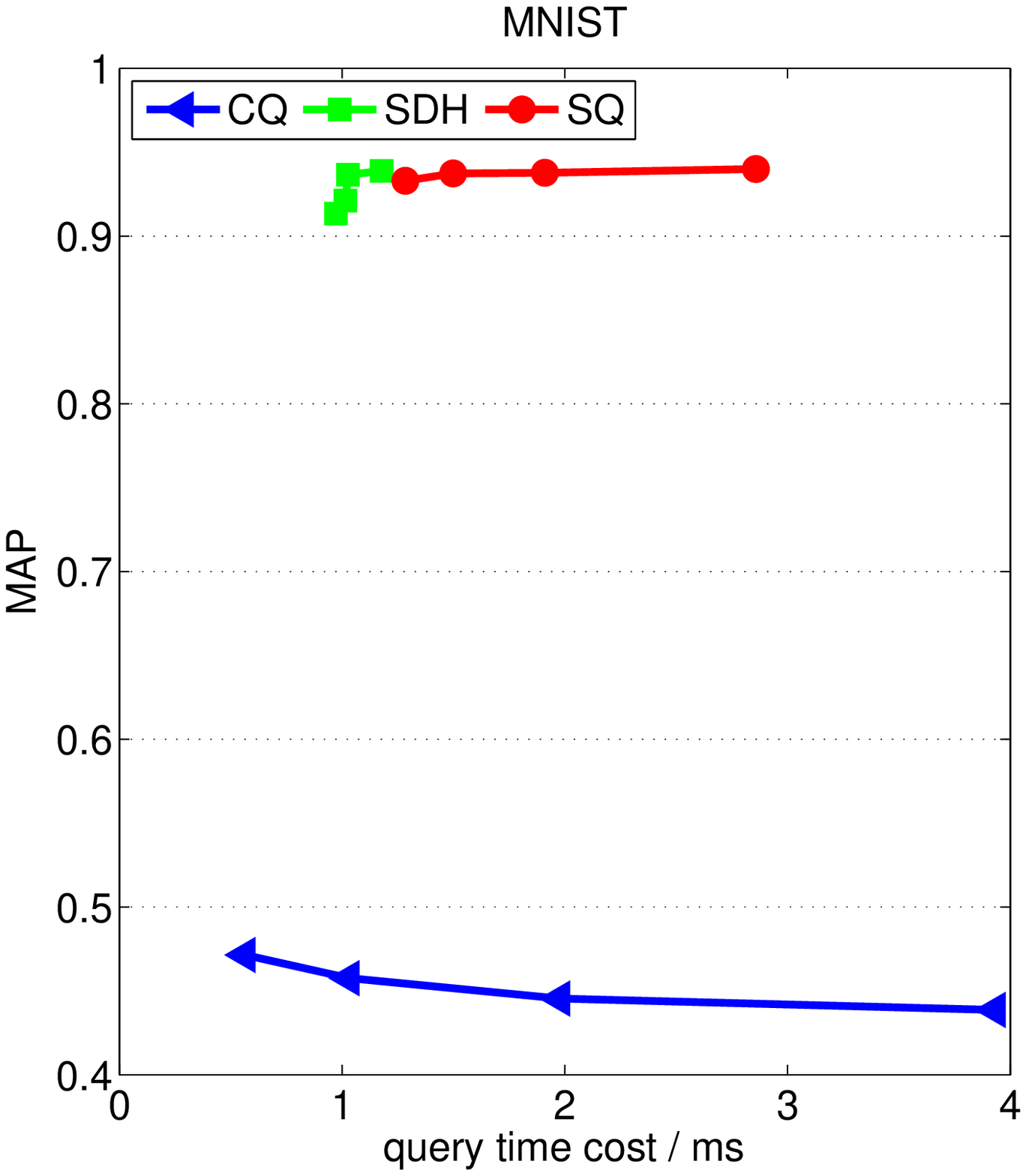}}
	\subfigure[NUS-WIDE]{\label{figure:querytime_nuswide}\includegraphics[width=.24\linewidth, clip]{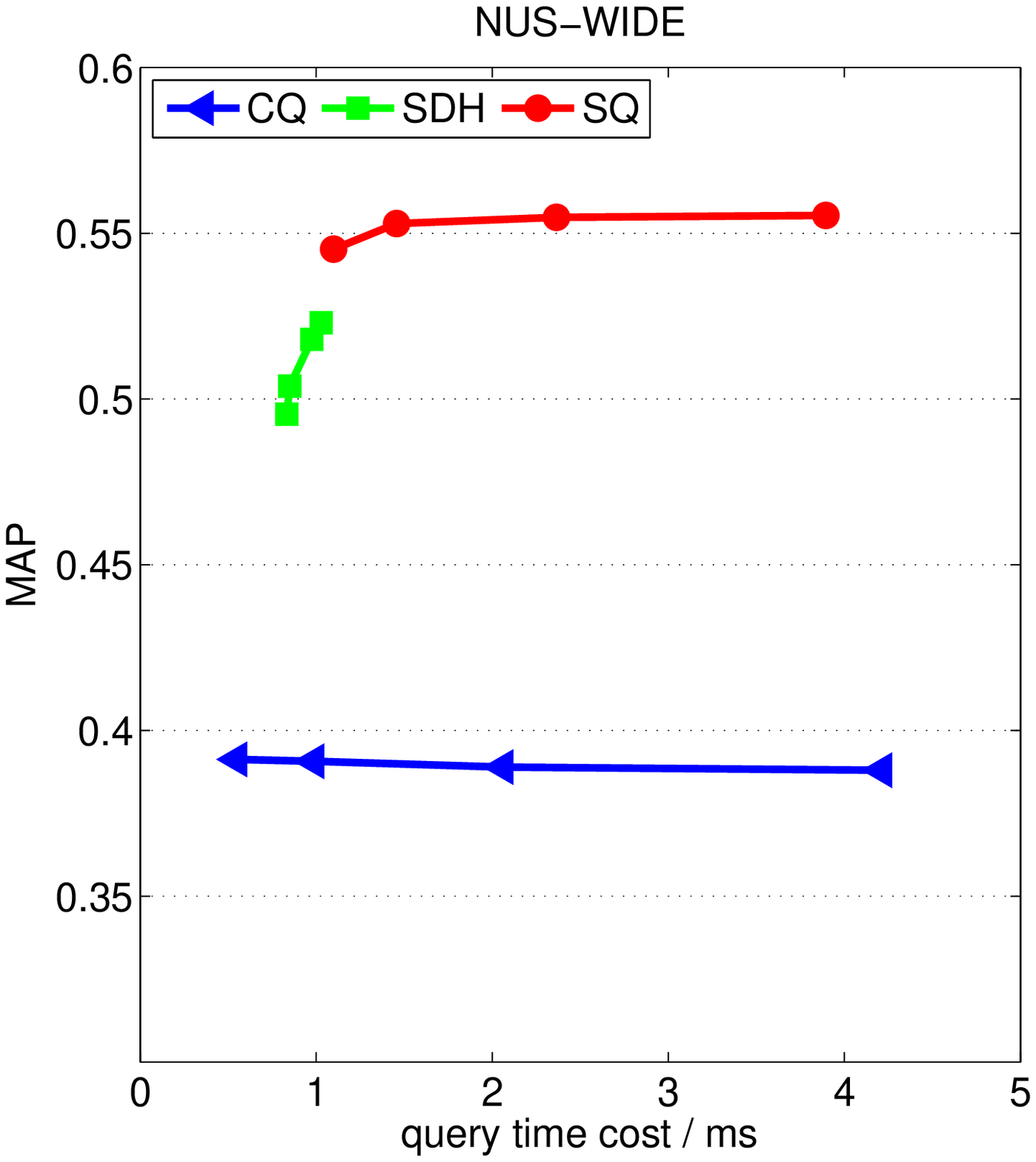}}
	\subfigure[ImageNet]{\label{figure:querytime_imagenet}\includegraphics[width=.24\linewidth, clip]{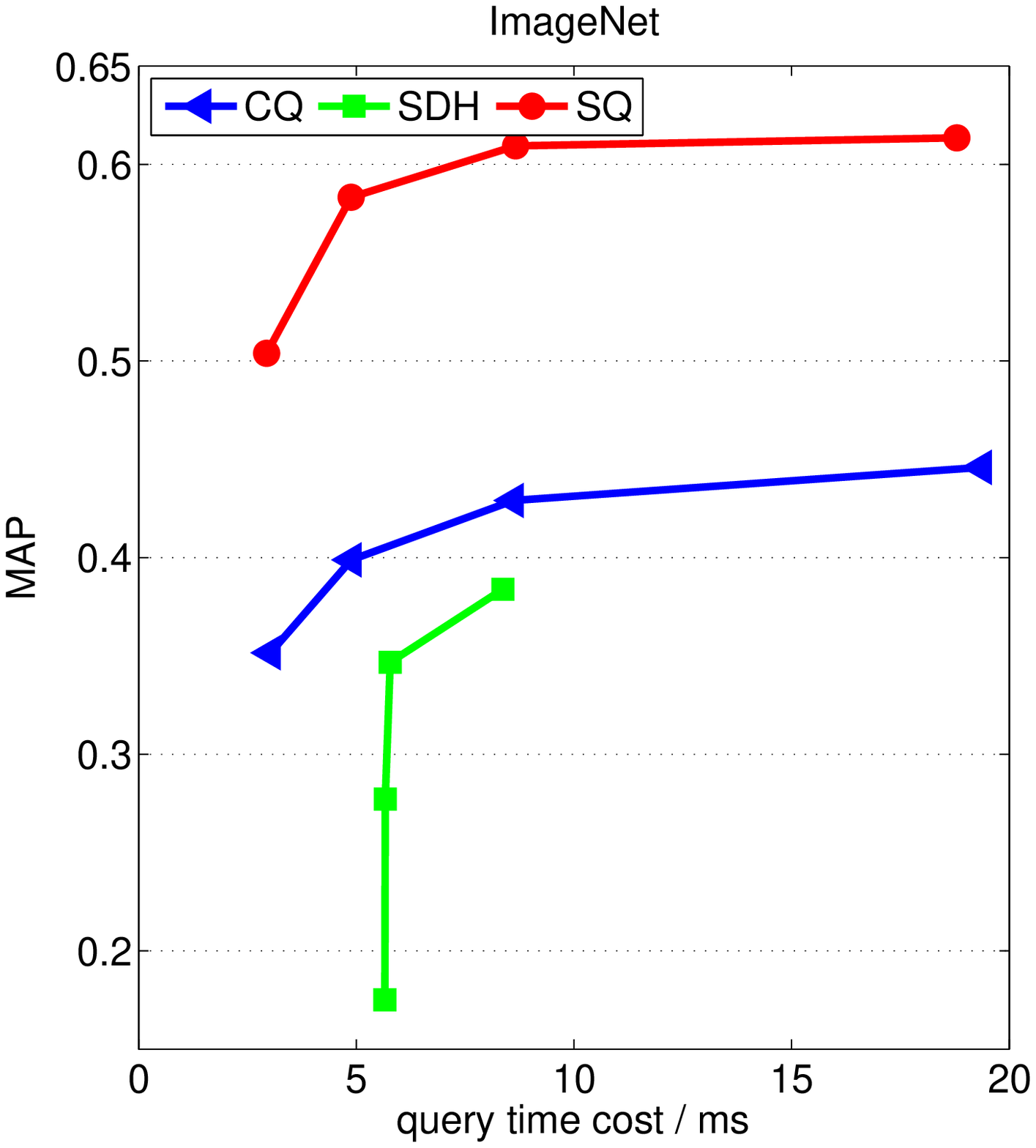}}
	\caption{Query time comparison of SQ, CQ, and SDH under various code lengths on CIFAR-10, MNIST, NUS-WIDE, and ImageNet.  The vertical axis represents the search performance, and the horizontal axis corresponds to the query time cost (milliseconds). The markers from left to right on each curve indicate the code length of $16$, $32$, $64$, and $128$ respectively. }
	\label{figure:queryTime}
\vspace{-.5cm}
\end{figure*}

\vspace{0.2cm}
\noindent {\bf Search efficiency.} We report the query time of our proposed approach SQ, the unsupervised quantization method CQ, and the supervised hashing method SDH, which outperforms other supervised hashing algorithms in our experiments.  Figure~\ref{figure:queryTime} shows the search performance and the corresponding query time under the code length of $16$, $32$, $64$, and $128$ on the four datasets.

Compared with CQ, our proposed SQ obtains much higher search performance for the same query time. It can be seen that on CIFAR-10, MNIST, and NUS-WIDE, SQ takes more time than CQ under the code length of $16$ and $32$, and less time under the code length of $128$: SQ takes extra time to do feature transformation; the querying process, however, is carried out in a lower-dimensional transformed subspace, therefore the search efficiency is still comparable to CQ.  It can also be observed that SQ takes almost equal time as CQ on ImageNet. This is because CQ also takes time to do feature transformation here and the querying process is carried out in the $256$-dimensional PCA subspace (it is cost prohibitive to tune the parameter of CQ on  high-dimensional large-scale dataset).

Compared with SDH, SQ outperforms SDH for the same query time on ImageNet and NUS-WIDE. For example, SQ with $32$ bits outperforms SDH with $16$ bits by a margin of $40.82\%$ on ImageNet, and SQ with $16$ bits outperforms SDH with $128$ bits by a margin of $2\%$ on NUS-WIDE, while they take almost the same query time.

On CIFAR-10, SQ with $16$ bits outperforms SDH with $128$ bits by $12.4\%$ while taking slightly more time ($0.16$ milliseconds), and this trend indicates that for the same query time, SQ could also obtain higher performance. On MNIST, SQ achieves the same performance as SDH while taking slightly more query time. The reason is that the query preprocessing time of SQ (mainly refers to distance lookup table construction time here) is relatively long compared with the linear scan search time on the small-scale database. In real-word scenarios, retrieval tasks that require quantization solution usually are conducted on large-scale databases, and the scale usually is at least $200,000$.

\subsection{Empirical analysis}\label{section::emperical analysis}
\begin{table}[t]
	\small
	\begin{center}
		\caption{MAP comparison of classification loss (denoted by ``c-loss") and triplet loss (denoted by ``t-loss").}
		\begin{tabular}{cccc cc}
			\hline
			Datasets & Methods        & 16 bits  & 32 bits   & 64 bits  & 128 bits\\ \hline
			\multirow{2}{*}{\footnotesize CIFAR-10}  & t-loss   & 0.3284 & 0.3679 & 0.5305   & 0.5469     \\
			& c-loss   & {\bf 0.6045}  & {\bf 0.6855} &{\bf 0.7042}   &{\bf 0.7120} \\
			\hline
			\multirow{2}{*}{\footnotesize MNIST}     & t-loss   & 0.4347  & 0.5286 & 0.6442 & 0.7500        \\
			& c-loss   & {\bf 0.9329} & {\bf 0.9374} & {\bf 0.9377} & {\bf 0.9400}  \\
			\hline
		\end{tabular}
		\label{table:: comparison with triplet loss}	
	\end{center}
\vspace{-.5cm}
\end{table}

\begin{figure*}[t]
	\centering
	\includegraphics[width=.48\linewidth, clip]{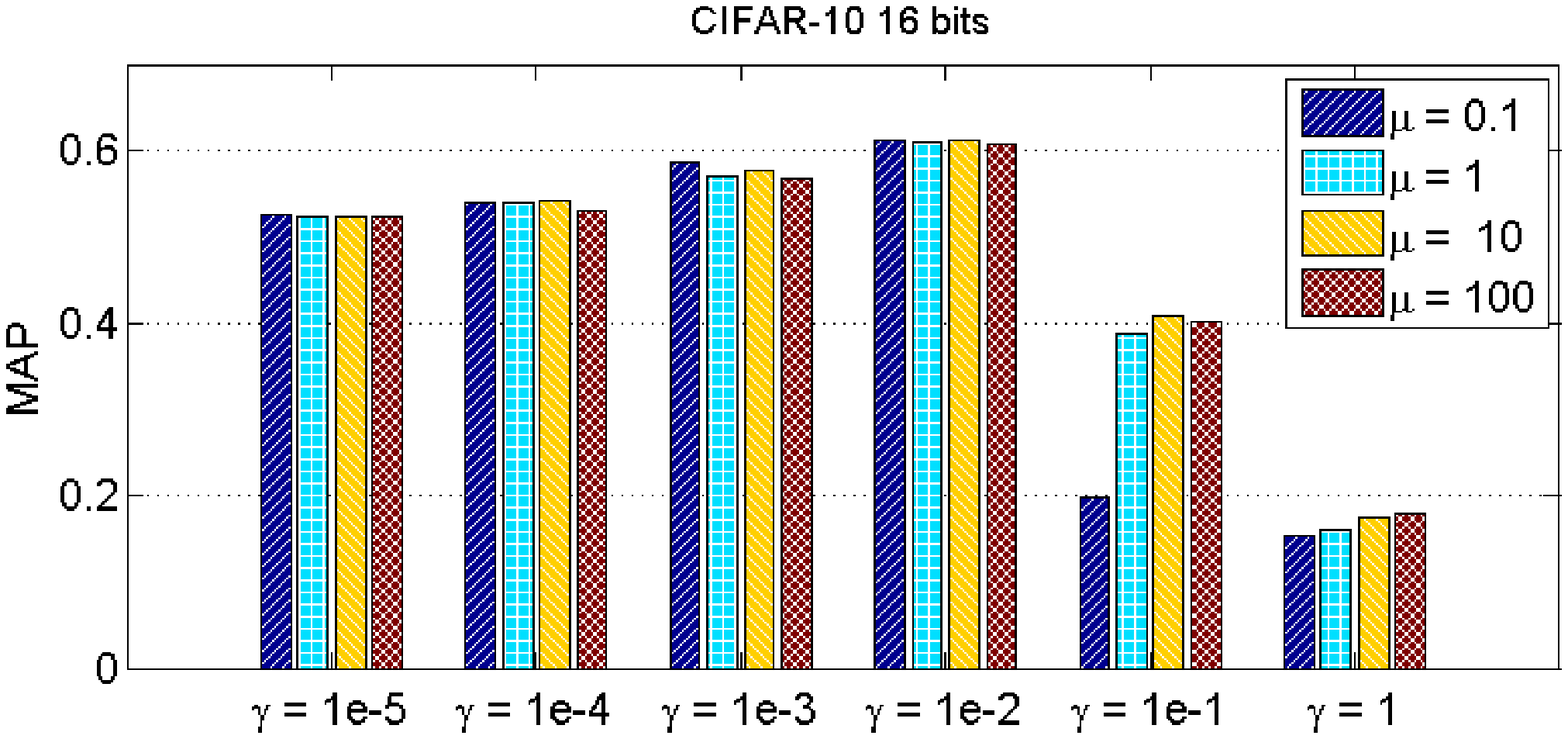}~
	\includegraphics[width=.48\linewidth, clip]{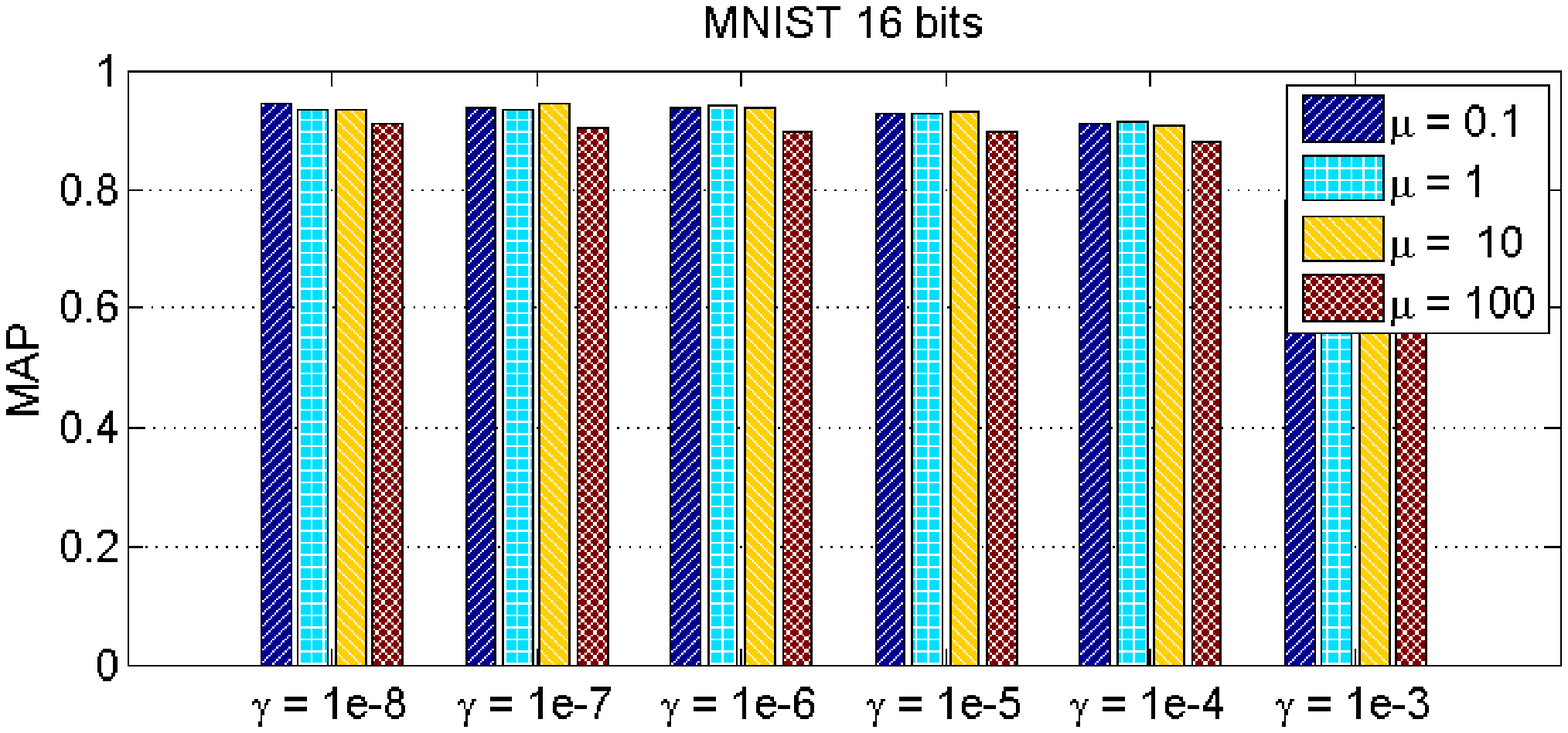}\\
	\includegraphics[width=.48\linewidth, clip]{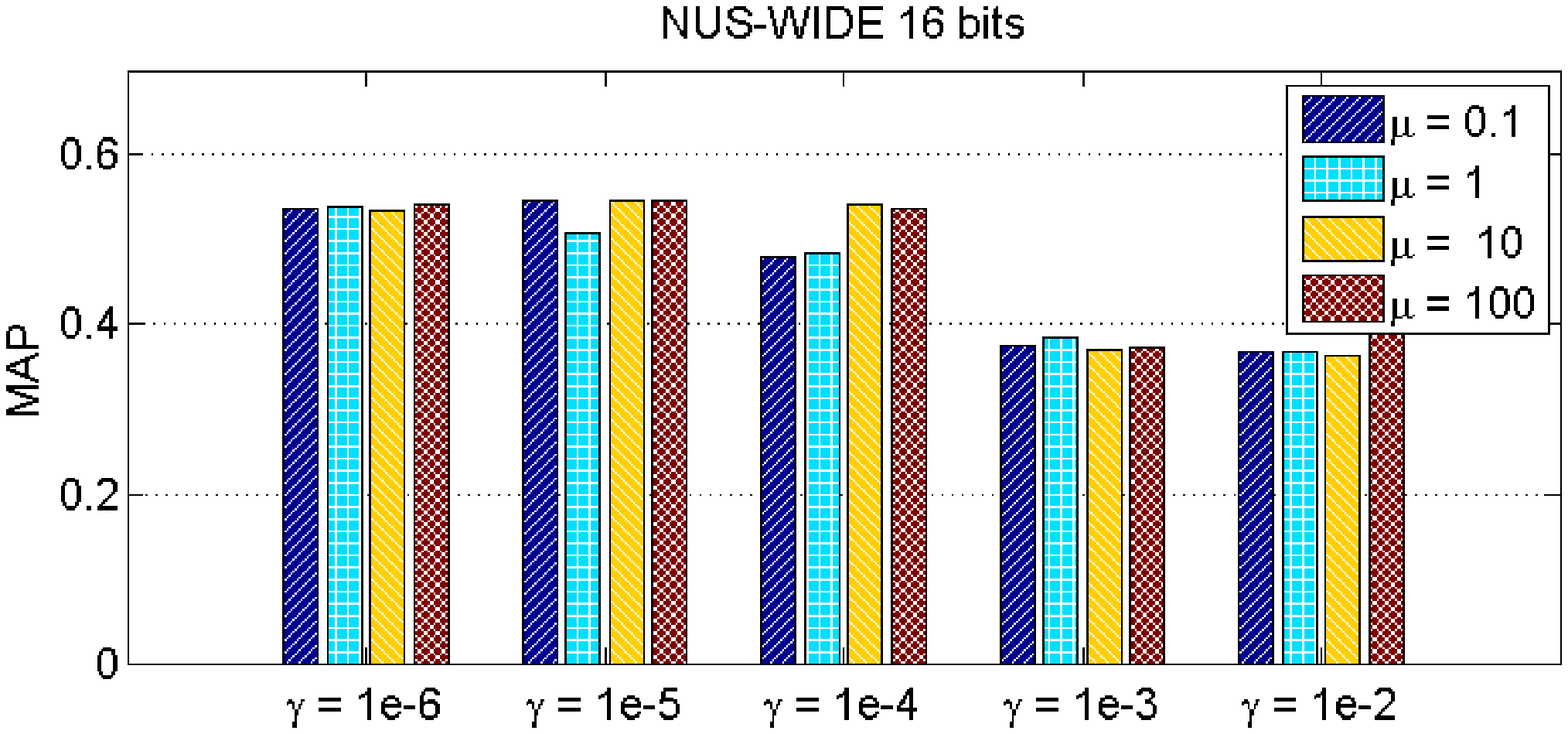}~
	\includegraphics[width=.48\linewidth, clip]{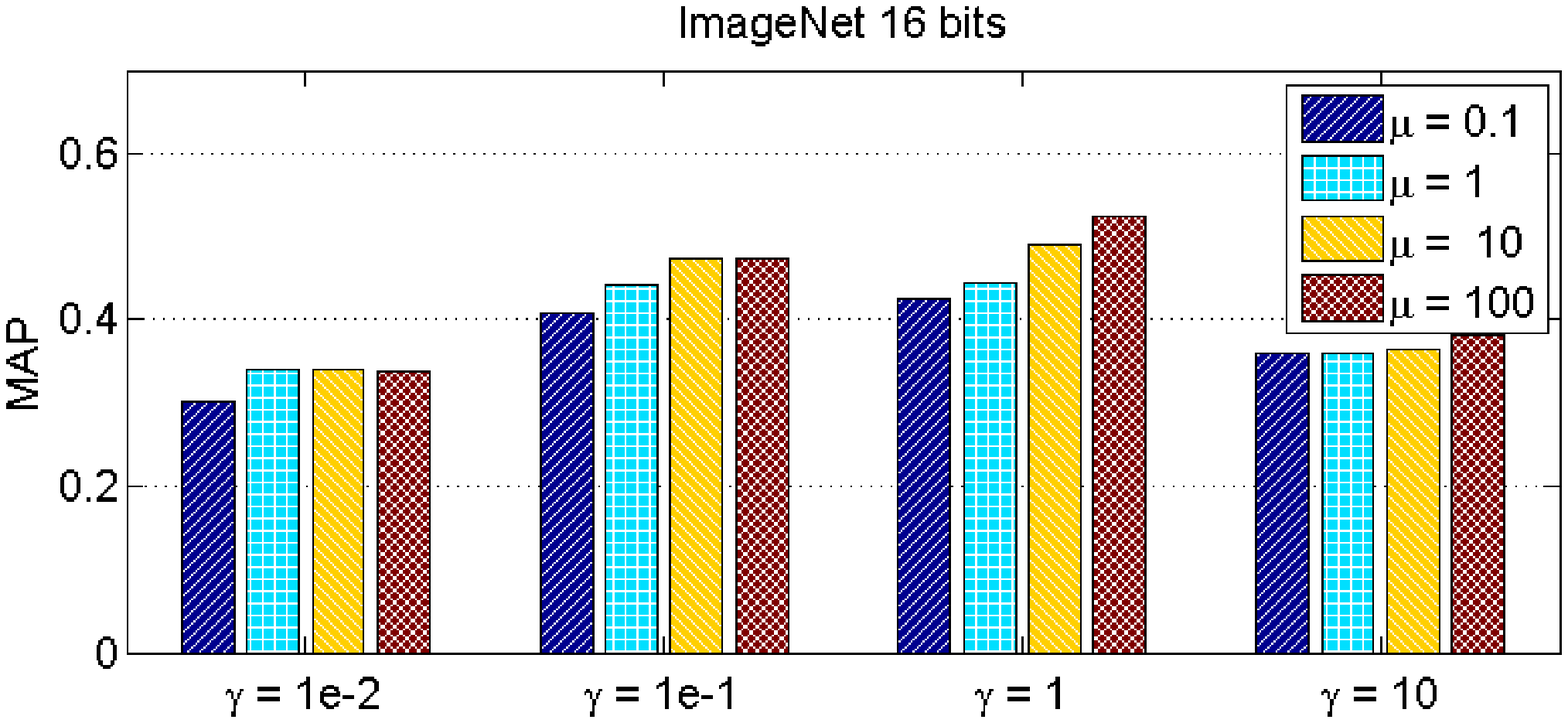}
	
	\caption{Illustration  of the effect of  $\gamma$ and $\mu$ on the search performance in the validation sets of CIFAR-10, MNIST, NUS-WIDE, and ImageNet with 16 bits. $\gamma$ ranges from 1e-7 to 1e+2
		and $\mu$ ranges from 1e-1 to 1e+2. }
	\label{figure::paramEval}
\vspace{-.5cm}
\end{figure*}

\noindent {\bf Classification loss vs. triplet loss.}
We empirically compare the performances
between
the proposed formulation~(\ref{equation::obj})
that uses the classification loss
for semantic separation,
and an intuitive formulation that uses triplet loss
to discriminate a semantically similar pair
and a semantically dissimilar pair. The triplet loss formulation is written as $ \sum_{(i, j, l)
	}[|| {\bf C}{\bf b}_i- {\bf C}{\bf b}_j||_2^2 - || {\bf C}{\bf b}_i- {\bf C}{\bf b}_l||_2^2+\rho]_+$. The triplet $(i, j , l)$ is composed of
three points
where $i$ and $j$ are from the same class
and $l$ is from a different class;
$\rho \ge 0$ is a constant indicating the distance margin;
$[\cdot]_+ = \max (0,\cdot)$ is the standard hinge loss function.

We optimize the formulation with triplet loss
using the alternative optimization algorithm
similar to that for optimizing problem~(\ref{equation::obj}).
The parameters $\gamma$ and $\mu$ are chosen through validation.
It is infeasible to do the optimization
with all the triplets.
Therefore we borrow the idea of active set,
and select the triplets that are most likely
to trigger the hinge loss at each iteration,
which is efficiently implemented
by maintaining an approximate nearest neighbor list for each database vector.

The results on CIFAR-10 and MNIST under various code lengths are shown in Table~\ref{table:: comparison with triplet loss}.
It is observed that the results with classification loss are much better than those with triplet loss.
It seems to us
that the triplet loss is better than classification loss,
as the search goal is essentially to rank similar pairs
before dissimilar pairs,
which is explicitly formulated in triplet loss.
The reason of the lower performance of triplet loss
most likely lies in the difficulty of the optimization 
(e.g., too many ($O(N^3)$) loss terms results in the sampling technique used for training, which makes the results not as good as expected).

\vspace{0.2cm}
\noindent{\bf Feature transformation.} Our approach learns the feature transformation matrix ${\bf P}$, and quantizes the database vectors in the learned discriminative subspace. To verify the effectiveness of feature transformation in our formulation~(\ref{equation::obj}), we empirically compare the performances between the proposed formulation and the  formulation that does not learn feature transformation. We take CIFAR-10 and MNIST as examples and the results are shown in Table~\ref{table:: comparison with no feature transformation}. As shown, SQ significantly outperforms the formulation that does not learn feature transformation, which indicates the importance of feature transformation in our proposed formulation.

\vspace{0.2cm}
\noindent {\bf The Effect of $\gamma$ and $\mu$.}\label{section:: discussion of paramters}
We empirically show how the parameters
$\gamma$ (for controlling the quantization loss term) and $\mu$ (for penalizing the  equality constraint term)
affect the search performance
on the validation set, where the parameters are tuned  to select the best combination.
We report the performances with $16$ bits
in Figure~\ref{figure::paramEval},
by varying $\gamma$ from 1e-7 to 1e+2
and $\mu$ from 1e-1 to 1e+2.

\begin{table}[t]
	\small
	\begin{center}
		\caption{MAP comparison of the formulation with feature transformation (denoted by ``with fea.") and that without feature transformation (denoted by ``no fea.").}
		\begin{tabular}{cccc cc}
			\hline
			Datasets & Methods        & 16 bits  & 32 bits   & 64 bits  & 128 bits\\ \hline
			\multirow{2}{*}{\footnotesize CIFAR-10} &  no fea.      & 0.5140 & 0.5174 & 0.5274   & 0.5301      \\
			& with fea.     & {\bf 0.6045}  & {\bf 0.6855} &{\bf 0.7042}   &{\bf 0.7120} \\
			\hline
			\multirow{2}{*}{\footnotesize MNIST}    & no fea.      & 0.4534 & 0.4538 & 0.4617	& 0.4650        \\
			& with fea.    & {\bf 0.9329} & {\bf 0.9374} & {\bf 0.9377} & {\bf 0.9400}  \\
			\hline
		\end{tabular}
		\label{table:: comparison with no feature transformation}	
	\end{center}
\vspace{-.6cm}
\end{table}

It can be seen from Figure~\ref{figure::paramEval}
that
the overall performances
do not depend much on $\mu$
and the performances change a lot
when varying the $\gamma$.
This is reasonable
because $\gamma$ controls the quantization loss,
and $\mu$ is introduced for accelerating the search.
The best search performances on CIFAR-10, MNIST, NUS-WIDE, and ImageNet
are obtained
with $(\gamma, \mu)$ = $(0.01, 0.1)$, $(\gamma, \mu)$ = $($1e-7$, 10)$,
$(\gamma, \mu)$ = $($1e-5$, 0.1)$, and $(\gamma, \mu)$ = $(1, 100)$ respectively.
We can see that the best MAP values $0.6132$,  $0.9449$, and $0.5466$ on the validation sets
are close to the values $0.6045$, $0.9329$, and $0.5452$ on the query sets of CIFAR-10, MNIST, and NUS-WIDE,
and that the MAP value $0.5372$ on the validation set is different from the value $0.5039$ on the query set of ImageNet. The reason might be that the validation set (sampled from the training set)
and the query set (the validation set provided in ImageNet) are not of the same distribution.

\section{Conclusion}
In this paper,
we present a supervised compact coding approach,
supervised quantization, to semantic similarity search. To the best of our knowledge,
our approach is the first attempt to study
the quantization for semantic similarity search.
The superior performance comes from
two points:
(\romannum{1}) The distance differentiation ability of quantization
is stronger than that of hashing. (\romannum{2}) The learned discriminative subspace
is helpful to find a semantic quantizer.

\section*{Acknowledgements}
This work was partially supported by the National Basic Research Program of China (973 Program) under Grant 2014CB347600.

{\small
\bibliographystyle{ieee}
\bibliography{ref}
}

\end{document}